\documentclass{bmvc2k}

\title{Learning a Neural Association Network for Self-supervised Multi-Object Tracking}

\addauthor{Shuai Li}{lishuai@iai.uni-bonn.de}{1}
\addauthor{Michael Burke}{michael.g.burke@monash.edu}{2,3}
\addauthor{Subramanian Ramamoorthy}{s.ramamoorthy@ed.ac.uk}{3}
\addauthor{Juergen Gall}{gall@iai.uni-bonn.de}{1,4}

\addinstitution{
 University of Bonn\\
 Bonn, Germany
}

\addinstitution{
 Monash University\\
 Melbourne, Australia
}

\addinstitution{
 University of Edinburgh\\
 Edinburgh, UK
}

\addinstitution{
 Lamarr Institute for Machine Learning and Artificial Intelligence\\
 Germany
}

\runninghead{Li, Burke, Ramamoorthy, Gall}{Neural Association}

\def\etal{\emph{et al}\bmvaOneDot}

\usepackage{cleveref}
\usepackage{amsmath,amsthm,amssymb}
\newcommand{\ie}{\textit{i}.\textit{e}.}

\usepackage{graphicx}
\newcommand{\scalemath}[2]{\scalebox{#1}{\(#2\)}}
\usepackage{booktabs}
\usepackage{algorithm}
\usepackage{multicol}
\usepackage{multirow}
\usepackage[noend]{algpseudocode}
\usepackage{bbding}
\usepackage{booktabs,makecell}

\begin{document}

\maketitle

\begin{abstract}
This paper introduces a novel framework to learn data association for multi-object tracking in a self-supervised manner. Fully-supervised learning methods are known to achieve excellent tracking performances, but acquiring identity-level annotations is tedious and time-consuming. Motivated by the fact that in real-world scenarios object motion can be usually represented by a Markov process, we present a novel expectation maximization (EM) algorithm that trains a neural network to associate detections for tracking, without requiring prior knowledge of their temporal correspondences. 
At the core of our method lies a neural Kalman filter, with an observation model conditioned on associations of detections parameterized by a neural network. Given a batch of frames as input, data associations between detections from adjacent frames are predicted by a neural network followed by a Sinkhorn normalization that determines the assignment probabilities of detections to states. Kalman smoothing is then used to obtain the marginal probability of observations given the inferred states, producing a training objective to maximize this marginal probability using gradient descent. 
The proposed framework is fully differentiable, allowing the underlying neural model to be trained end-to-end. We evaluate our approach on the challenging MOT17, MOT20, and BDD100K datasets and achieve state-of-the-art results in comparison to self-supervised trackers using public detections. 
\end{abstract}

\section{Introduction}
\label{sec:intro}
Multi-object tracking (MOT) is highly relevant for many applications ranging from autonomous driving to understanding the behavior of animals. Thanks to the rapid development of object detection algorithms~\cite{felzenszwalb2009object, ren2015faster,yang2016exploit}, tracking-by-detection~\cite{kim2015multiple, hornakova2020lifted, li2022learning} has become the dominant paradigm for multi-object tracking. Given an input video, a set of detection hypotheses is first produced for each frame and the goal of tracking is to link these detection hypotheses across time, in order to generate plausible motion trajectories.

A large number of learning-based methods have focused on the fully-supervised MOT setting~\cite{hornakova2020lifted,li2022learning,cetintas2023unifying}. These approaches assume that a training set with detections together with the
associations between these detections are provided, and the goal is to train a model that can predict data association between detections during inference. While these approaches achieve strong performance on standard tracking benchmarks~\cite{milan2016mot16, dendorfer2019cvpr19}, they demand costly labeling and annotation burdens that can be expensive to scale. In contrast to fully-supervised methods, self-supervised approaches seek to train a model that is able to temporally associate noisy detections, without requiring knowledge of the data association between them during training. Following this trend, Favyen~\etal~\cite{bastani2021self} designed a method that takes two modalities as input. While the first contains only bounding-box coordinates, the second contains appearance information solely. During training, the two inputs are forced to output the same association results. Similarly, Lu~\etal~\cite{lu2024self} suggest to drop several detections within a track and utilize a path consistency constraint to train a network for association. This method achieves state-of-the-art performance in the self-supervised setting, but requires a number of heuristics and involves a complex removal strategy for training, which makes the training very expensive. 

\begin{figure}[!t]
    \centering
    \begin{minipage}{0.6\linewidth}
        \centering
        \includegraphics[width=\linewidth]{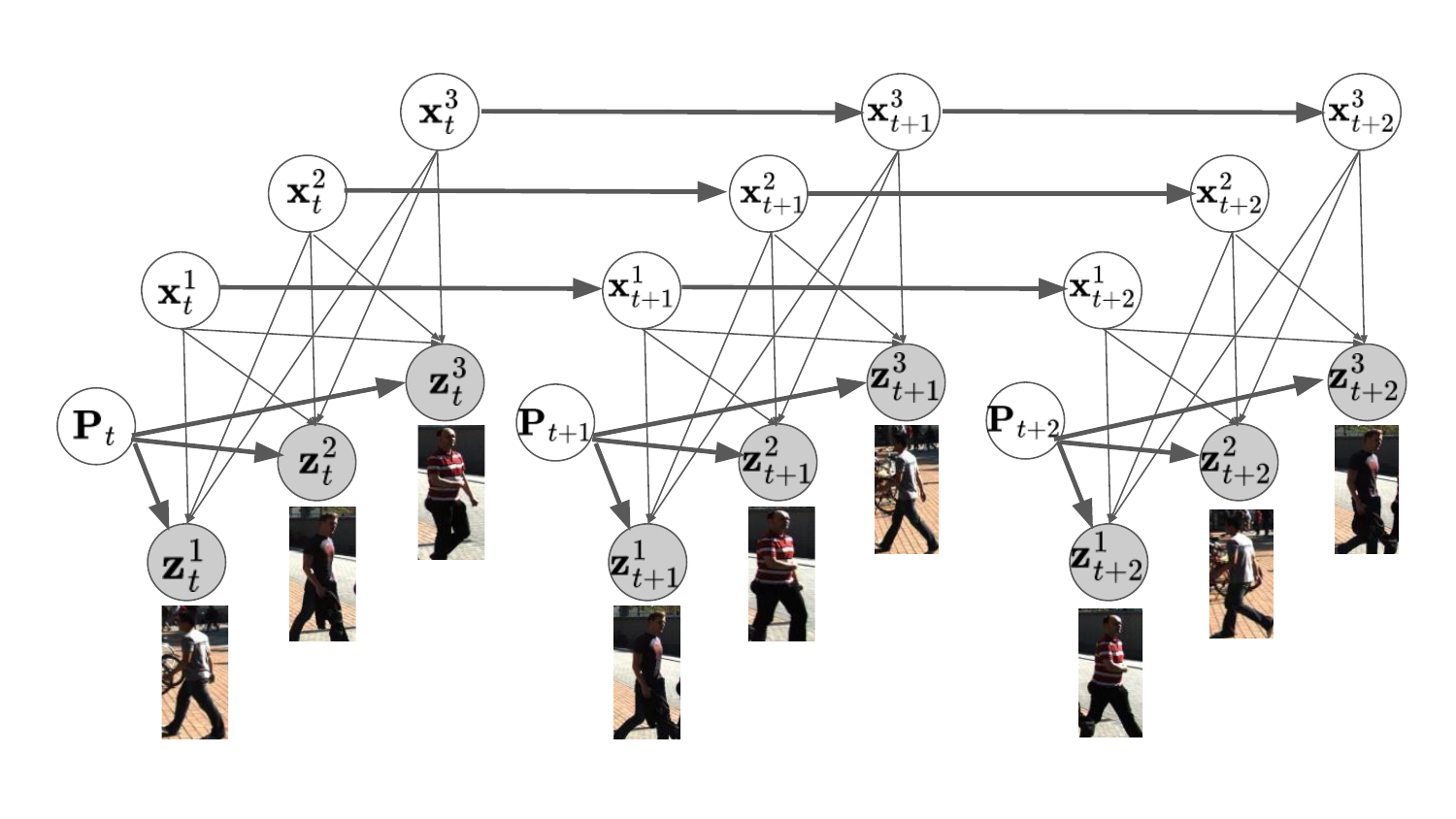}
    \end{minipage}\hfill
    \begin{minipage}{0.35\linewidth}
        \caption{Given observations $\mathbf{z}_{1:T}$, a neural network is trained to predict the permutations $\mathbf{P}_{1:T}$ that assign observations to states $\mathbf{x}_{1:T}$, which define tracks. The network is trained self-supervised, \ie, without given associations between $\mathbf{z}_{1:T}$ and $\mathbf{x}_{1:T}$. After training, the network is used to associate detections in MOT.}
        \label{figure:teaser}
    \end{minipage}
\end{figure}

In this work, we introduce a novel framework to learn data association in a self-supervised manner. Motivated by the fact that object motion can be usually represented by a smooth Markov process, we propose an Expectation Maximization (EM) algorithm that finds associations, which rewards locally smooth trajectories over non-smooth associations. The core of our method relies on a neural Kalman filter~\cite{kalmanfilter, krishnan2015deep} and a differentiable assignment mechanism. The Kalman filter provides a principled way to model uncertainty in an efficient way since densities are evaluated in closed form. In particular, our Kalman filter's motion model is realized by a random walk process while the observation model is parameterized by a neural network that provides the assignment probabilities of the observations to states via a permutation matrix. In other words, the permutation matrix associates detections in a video to the corresponding tracks, as illustrated in Fig.~\ref{figure:teaser}. Since the permutation matrix should be doubly-stochastic, we propose to add a Sinkhorn layer into the system. Kalman smoothing~\cite{rauch1965maximum} is then used to obtain the marginal probability of observations given inferred state trajectories, leading to a training objective that maximizes this marginal. As the Sinkhorn iterations merely involve normalization across rows and columns for several steps, the entire procedure is fully differentiable, allowing the underlying neural model to be trained in an end-to-end manner using gradient descent.

We further show how this approach can be used to successfully fine-tune an appearance model on the MOT17~\cite{milan2016mot16} training set using association results produced by the trained association model, such that it enables better association abilities. During inference, we follow an online Kalman filter tracking paradigm~\cite{wojke2017simple} where the detections within the current frame are matched to the predicted detections from the previous frame using the learned association network.
 
In comparison to prior works~\cite{bastani2021self, lu2024self}, our approach provides a principled and probabilistic way for self-supervised learning. The model can be trained in just a few minutes, which is much faster than~\cite{bastani2021self, lu2024self}, which  require about 24 hours for training. We evaluate our approach on the MOT17, MOT20~\cite{milan2016mot16,dendorfer2019cvpr19}, and BDD100K~\cite{yu2020bdd100k} datasets, and we show that our approach achieves better or comparable results than existing self-supervised multi-object tracking approaches. 
Our contributions are summarized as follows:
\begin{itemize}
\item We propose a novel self-supervised framework that embraces the Expectation Maximization algorithm to learn data association for MOT.
\item The framework enables us to learn motion as well as appearance affinity together for robust data association and the learned association network naturally fits into the online tracking paradigm.
\item Our approach achieves state-of-the-art performance among existing self-supervised MOT methods on the challenging MOT17 and MOT20 datasets with public detections and on the BDD100K~\cite{yu2020bdd100k} dataset.
\end{itemize}

\section{Related Work}
\noindent{\textbf{Self-Supervised Multi-Object Tracking.}}
Self-supervised MOT has the advantage that the training of data association models does not require expensive identity-level supervision, compared to its fully-supervised counterpart. SORT \cite{bewley2016simple} adopts a simple Intersection-over-Union (IoU) as an affinity metric for data association. This approach, however, is sensitive to occlusions. Following this trend, Yang~\etal~\cite{yang2023hard} present a heuristic buffered IoU (BIoU) metric for data association. Karthik~\etal~\cite{karthik2020simple} utilize SORT~\cite{bewley2016simple} to generate pseudo track labels for training an appearance model which is then used in an online tracking framework. Favyen~\etal~\cite{bastani2021self} propose an interesting input masking strategy that enforces mutual consistency between two given input modalities as an objective during training, an approach that achieves decent results for data association. Lin~\etal~\cite{lin2022unsupervised} present a dynamical recurrent variational autoencoder architecture, but this approach requires pre-training of motion models, while the data association between states and observations admits a simple closed-form solution, it can only track a fixed number of objects. An interesting recent work is PKF~\cite{cao2024pkf}, which assigns observations to a given track in a soft, probabilistic way, at the sacrifice of MOTA metirc. In contrast to these works, our approach provides a principled objective and it requires only minutes to train.

\section{Neural Data Association using Expectation Maximisation}
In this work, we propose a self-supervised learning approach for multi-object tracking. This means that only a set of unlabeled detections in a batch of frames are given and the goal is to train a neural network such that it can predict the associations of these detections accurately. 

We rephrase this problem as a Kalman filtering problem as shown in Fig.~\ref{figure:teaser}. The unknown states of $K$ objects at the $t$-th time step are represented as $\mathbf{x}_t \in \mathbb{R}^{K \times d}$. Each object is represented by its $d$-dimensional state vector, containing its $x, y$ central coordinates as well as its velocity in the image plane,~\ie $\mathbf{x} = (x, y, \dot{x}, \dot{y})$. The track of an object $k$ over $T$ frames is thus indicated by $\mathbf{x}^k_{1:T}$. $\mathbf{z}_t \in \mathbb{R}^{K \times d^\prime}$ represents the observed detections where each detection at frame $t$ is represented by a $d'$-dimensional observation vector, containing bounding box coordinates and appearance features. In contrast to a classical Kalman filtering problem, we do not know to which state $\mathbf{x}^k_{t}$ an observation $\mathbf{z}^l_t$ belongs to. This association is modeled by a permutation matrix $\mathbf{P}_t$ which assigns the observations $\mathbf{z}_{t}$ to the states $\mathbf{x}_{t}$. For training, we assume that all $K$ objects are visible in all $T$ frames and we will discuss occlusions or false positive detections later.

Given $\mathbf{P}_t$, we can update the states using a linear Kalman filter~\cite{kalmanfilter} with Gaussian noise:
%
\begin{align} \label{eq:kalman_motion_equation}
p(\mathbf{x}_t | \mathbf{x}_{t-1}) &= \mathcal{N}(\mathbf{x}_t; \mathbf{Fx}_{t-1}, \mathbf{Q}_t),\\
\label{eq:kalman_observation_equation}
p(\mathbf{z}_t | \mathbf{x}_t, \mathbf{P}_t) &= \mathcal{N}(\mathbf{z}_t; \mathbf{H}_t\mathbf{P}_t\mathbf{x}_t, \mathbf{R}_t),
\end{align}
%
where $\mathbf{Q}_t$ and $\mathbf{R}_t$ are Gaussian noise, $\mathbf{H}_t$ is the linear mapping from state space to observation space, and $\mathbf{x}_t = \mathbf{Fx}_{t-1}$ is the motion model. While we use a linear model, the approach can be extended to non-linear models or a learned dynamic model. $\mathbf{P}_t$ will be estimated by a neural network $g_{\theta}$ and we will learn the parameters of the network using expectation maximisation and back propagation.

\begin{figure}[!t]
\centering
\includegraphics[width=0.8\textwidth]{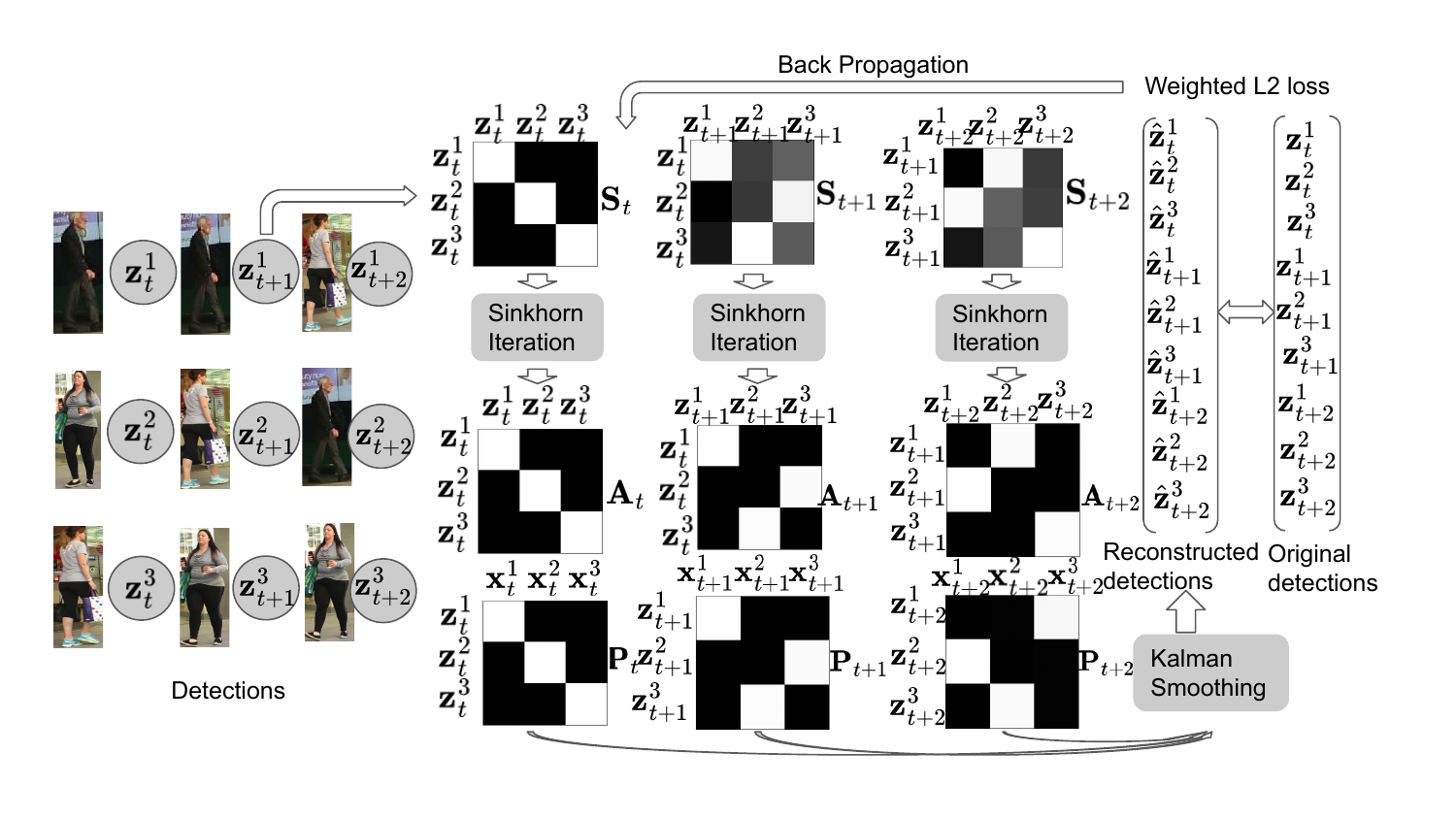}
\vspace{-0.5cm}
\caption{The proposed self-supervised learning framework. Given \textit{unlabeled} detections $\mathbf{z}_{t:t+2}$, we train a neural network $g_{\theta}$ that predicts the score matrices $\mathbf{S}_{t:t+2}$ between detections from every pair of adjacent frames and for the first frame between all detections $\mathbf{z}_{t}$ of the same frame. For $\mathbf{S}_{t}$, the values should thus be high on the diagonal but low everywhere else. Sinkhorn normalization is then applied to $\mathbf{S}_{t:t+2}$ to produce the doubly stochastic association matrix $\mathbf{A}_{t:t+2}$, which associates detections between frames. The permutation matrices $\mathbf{P}_{t:t+2}$ then assign the detections to the states $\mathbf{x}$, where the state $\mathbf{x}^k_{t:t+2}$ represents a track of an object $k$. Kalman smoothing combined with the predicted $\mathbf{P}_{t:t+2}$ then reconstructs the detections, which are compared with original detections to define the loss in~\cref{eq:kalman_smoothing_objective}. This loss is then back-propagated to update the parameters of the network $g_{\theta}$. We only show three frames for simplicity, but the framework works on longer sequences.}
\label{figure:pipeline}
\end{figure}

\subsection{Maximum Likelihood Learning Framework}

Given the linear Gaussian motion and observation models (\cref{eq:kalman_motion_equation,eq:kalman_observation_equation}), we can compute a predictive posterior, represented by state's mean $\hat{\boldsymbol{\mu}}_t$ and covariance $\hat{\boldsymbol{\Sigma}}_t$ at timestep $t$, in closed form using a Kalman filter prediction step conditioned on a series of permutations $\mathbf{P}_{1:t-1}$:
\vspace{-0.3cm}
\begin{align} \label{eq:kalman_prediction}
\nonumber p(\mathbf{x}_t|\mathbf{z}_{1:t-1}, \mathbf{P}_{1:t-1})=& \int p(\mathbf{x}_t|\mathbf{x}_{t-1})p(\mathbf{x}_{t-1}|\mathbf{z}_{1:t-1}, \mathbf{P}_{1:t-1})d\mathbf{x}_{t-1}\\
\nonumber =& \mathcal{N}(\mathbf{x}_t; \mathbf{F}\boldsymbol{\mu}_{t-1}, \mathbf{F}\boldsymbol{\Sigma}_{t-1}\mathbf{F}^T + \mathbf{Q}_t)\\
 =& \mathcal{N}(\mathbf{x}_t; \hat{\boldsymbol{\mu}}_t, \hat{\boldsymbol{\Sigma}}_t)\,.
\end{align}

Here, $p(\mathbf{x}_{t-1}|\mathbf{z}_{1:t-1}, \mathbf{P}_{1:t-1})$ is the Gaussian posterior at time $t-1$ from the last Kalman filter update step. Once the observations $\mathbf{z}_t$ are available at the current timestamp $t$, the update equation for the posterior is given by:
%
\begin{align} \label{eq:kalman_update}
p(\mathbf{x}_t|\mathbf{z}_{1:t}, \mathbf{P}_{1:t}) =& \frac{p(\mathbf{z}_t|\mathbf{x}_t, \mathbf{P}_t) p(\mathbf{x}_t|\mathbf{z}_{1:t-1}, \mathbf{P}_{1:t-1})}{p(\mathbf{z}_t|\mathbf{z}_{1:t-1}\mathbf{P}_{1:t})}\\
\nonumber =& \mathcal{N}(\mathbf{x}_t; \boldsymbol{\mu}_t, \boldsymbol{\Sigma}_t)\\
\label{eq:kalman_marginal}p(\mathbf{z}_t|\mathbf{z}_{1:t-1}, \mathbf{P}_{1:t}) =& \int p(\mathbf{z}_t|\mathbf{x}_t, \mathbf{P}_t) p(\mathbf{x}_t|\mathbf{z}_{1:t-1}, \mathbf{P}_{1:t-1}) d\mathbf{x}_t\\
\nonumber =& \mathcal{N}(\mathbf{z}_t; \mathbf{H}_t \mathbf{P}_t \hat{\boldsymbol{\mu}}_t, (\mathbf{H}_t \mathbf{P}_t)\hat{\boldsymbol{\Sigma}}_t (\mathbf{H}_t \mathbf{P}_t)^T + \mathbf{R}_t).
\end{align}

The derivation above uses a Kalman filter that makes predictions based on a history of observations. During training, we use Kalman smoothing~\cite{rauch1965maximum} to calculate the marginal as in \cref{eq:kalman_smoothing_marginal}, where $\tilde{\boldsymbol{\mu}}_t$ and $\tilde{\boldsymbol{\Sigma}}_t$ denote the smoothed mean and covariance of the current state $\mathbf{x}_t$, respectively: 
\vspace{-0.3cm}
\begin{align} \label{eq:kalman_smoothing_marginal}
p(\mathbf{z}_t|\mathbf{z}_{1:T}, \mathbf{P}_{1:T}) =& \int p(\mathbf{z}_t, \mathbf{x}_t| \mathbf{z}_{1:T}, \mathbf{P}_{1:T}) d\mathbf{x}_t\\
\nonumber =& \int p(\mathbf{z}_t| \mathbf{x}_t, \mathbf{z}_{1:T}, \mathbf{P}_{1:T}) p(\mathbf{x}_t | \mathbf{z}_{1:T}, \mathbf{P}_{1:T}) d\mathbf{x}_t\\
\nonumber =& \int \mathcal{N}(\mathbf{z}_t; \mathbf{H}_t\mathbf{P}_t\mathbf{x}_t, \mathbf{R}_t)\mathcal{N}(\mathbf{x}_t; \tilde{\boldsymbol{\mu}}_t, \tilde{\boldsymbol{\Sigma}}_t) d\mathbf{x}_t\\
\nonumber =& \mathcal{N}(\mathbf{z}_t; \mathbf{H}_t \mathbf{P}_t \tilde{\boldsymbol{\mu}}_t, (\mathbf{H}_t \mathbf{P}_t)\tilde{\boldsymbol{\Sigma}}_t (\mathbf{H}_t \mathbf{P}_t)^T + \mathbf{R}_t).
\end{align}

This uses both a forward and backward process to condition on all observations $\mathbf{z}_{1:T}$, rather than conditioning only on prior observations at a given time step $\mathbf{z}_{1:t}$. It not only reduces uncertainty but also helps to ensure forward and backward temporal consistency in the associations of $\mathbf{z}_{1:T}$, providing more robust training. 

\noindent{\textbf{Estimating $\mathbf{P}_{1:T}$.}}
We delineate the detailed implementation pipeline in Fig.~\ref{figure:pipeline}. Given detections $\mathbf{z}_{t-1}^i$ and $\mathbf{z}_{t}^j$ from the frames $t-1$ and $t$, respectively, a network predicts their similarity $s_{ij}=g_{\theta}(\mathbf{z}_{t-1}^i,\mathbf{z}_{t}^j)$, with a higher value of $s_{ij}$ indicating a higher similarity. By iterating over all detection pairs, we obtain a score matrix $\mathbf{S}_{t}$. From the score matrix, the Sinkhorn layer computes the association matrix $\mathbf{A}_{t}$ that associates detections between adjacent frames. As the permutation matrix $\mathbf{P}_t$ is required for training, we initialize the states in the first frame using the observations from the first frame, \ie, $\mathbf{x}_1=\mathbf{z}_1$, making $\mathbf{A}_1$ as the identity matrix. $\mathbf{P}_t$ is then obtained by $\mathbf{P}_t = \prod_{i=t}^{1} \mathbf{A}_i$.

\noindent{\textbf{Sinkhorn Layer.}}
In general, the permutation $\mathbf{P}$ is a hard assignment matrix that makes it non-trivial to back-propagate the loss through it. It is therefore desirable to make it a ``soft'' version to enable gradient-based end-to-end training. The permutation matrix $\mathbf{P}$, however, needs to be doubly stochastic, \ie, its rows and columns should sum to one. To address this, we propose to add a Sinkhorn layer~\cite{sinkhorn1964relationship,mena2018learning} to normalize the score matrix $\mathbf{S}$:
%
%
\begin{equation}
    \label{eq:Sinkhorn}
    \scalemath{0.9}
    {
    X^0(\mathbf{S}) = \exp(\mathbf{S}), \quad
    X^l(\mathbf{S}) = \mathcal{T}_{col}(\mathcal{T}_{row}(X^{l-1}(\mathbf{S}))), \quad
    \mathbf{A} = \lim_{l \rightarrow \infty} X^l(\mathbf{S}).
    }
\end{equation}

It applies the Sinkhorn operator~\cite{sinkhorn1964relationship} $X$ to the square matrix $\mathbf{S}$. In particular, $\mathcal{T}_{row}(\mathord{\cdot})$ and $\mathcal{T}_{col}(\mathord{\cdot})$ indicate row-wise and column-wise normalization operations, respectively. Repeating this for several iterations creates a doubly stochastic data association matrix $\mathbf{A}$ between detections. By combining~\cref{eq:kalman_smoothing_marginal} and $\mathbf{P}_{1:T}$, which depends on $\mathbf{S}_{1:T}$ and thus $g_{\theta}$, we define our training objective for $\theta$, which is optimized by gradient descent:
%
\begin{align} \label{eq:kalman_smoothing_objective}
\scalemath{0.9}{
\mathop{\arg \min}_{\theta} -\sum_{t=1}^T \log \mathcal{N}(\mathbf{z}_t; \mathbf{H}_t \mathbf{P}_t \tilde{\boldsymbol{\mu}}_t, (\mathbf{H}_t \mathbf{P}_t)\tilde{\boldsymbol{\Sigma}}_t (\mathbf{H}_t \mathbf{P}_t)^T + \mathbf{R}_t).
}
\end{align}

The detailed training algorithm is shown in~\Cref{alg1}. Intuitively, this can be seen as an expectation maximisation (EM) approach that alternates between inferring underlying state trajectories, and identifying permutations mapping observations to trajectories.

\begin{algorithm}[!t]
\small
\caption{The proposed procedure for learning data association.}
\label{alg1}
\hspace*{0.02in} 
{\bf Input:} Observations $\mathbf{z}_{1:T}$, learning rate $\alpha$\\
\hspace*{0.02in} 
{\bf Output:} $g_{\theta}(\mathord{\cdot})$

\begin{algorithmic}[1]
\State {Initialize $\mathbf{A}_1$ with identity matrix, MLP with $\mathbf{\theta}_{0}$} 
\For{$n=1$ to number of iterations~$N$}
\For{$t=2$ to $T$}
\State $\mathbf{S}_t = g_{\theta}(\mathbf{z}_{t-1}, \mathbf{z}_t)$ 
\State Predict $\mathbf{A}_t$ using Sinkhorn iteration~(\cref{eq:Sinkhorn})
\State $\mathbf{P}_t = \prod_{i=t}^1 \mathbf{A}_i$
\State Compute $p(\mathbf{x}_t|\mathbf{z}_{1:t-1}, \mathbf{P}_{1:t-1})$ using~\cref{eq:kalman_prediction}
\State Compute $p(\mathbf{x}_t|\mathbf{z}_{1:t}, \mathbf{P}_{1:t})$ using~\cref{eq:kalman_update}
\EndFor
\For{$t=T$ to $1$}
\State Compute $p(\mathbf{z}_t|\mathbf{z}_{1:T}, \mathbf{P}_{1:T})$ using~\cref{eq:kalman_smoothing_marginal}
\EndFor
\State Compute $\mathcal{L}{=}-\sum_{t=1}^T \log p(\mathbf{z}_t|\mathbf{z}_{1:T}, \mathbf{P}_{1:T})$ 
\State $\mathbf{\theta}_{n+1} = \mathbf{\theta}_n - \alpha  \frac{\partial \mathcal{L}}{\partial \mathbf{\theta}}$
\EndFor

\end{algorithmic}
{\bf Return} Learned MLP $g_{\theta_N}(\mathord{\cdot})$
\end{algorithm}

\noindent{\textbf{Learning the Appearance Model.}}
The aforementioned procedure only involves learning association between detections from adjacent frames using relative geometrical features. However, it is also desirable to learn an appearance model in order to associate objects when geometrical information is unreliable due to abrupt camera motion. To this end, we use the inferred associations from the previous step to finetune an appearance model $\phi_{\theta}(\mathord{\cdot})$ parameterized by the ImageNet pretrained ResNet-50~\cite{he2016deep}.

Given a training sample that contains $K \times T$ detections, we first calculate the permutation matrix $\mathbf{P}_T \in \mathbb{R}^{K \times K}$, such that the rows indicate detections at the $T$-th and the columns denote detections at the first frame. 
Each element $p_{ij}$ represents the probability that detection $i$ at frame $T$ is associated to the detection $j$ at frame 1. We also construct a similarity matrix $\mathbf{U}_T \in \mathbb{R}^{K \times K}$ using the cosine similarity of appearance features, \ie,  
\begin{equation}
u_{ij} = \frac{\phi_{\theta}(\mathbf{z}_T^i)^T\phi_{\theta}(\mathbf{z}_1^j)}{||\phi_{\theta}(\mathbf{z}_T^i)||^2||\phi_{\theta}(\mathbf{z}_1^j)||^2},    
\end{equation}
where $\mathbf{U}_T$ is normalized row-wise through softmax. 

Intuitively, if the $i$-th detection at frame $T$ is associated with the $j$-th detection at the first frame, then their appearance similarity should also be high. Since $\mathbf{P}_T$ is a soft permutation matrix, we propose to minimize the KL-divergence between $\mathbf{P}_T$ and $\mathbf{U}_T$ as a second loss:
\begin{equation}\label{eq:kl_divergence}
\scalemath{0.85}{
\displaystyle D_{KL}(\mathbf{P_T} || \mathbf{U_T}) = \sum_{i=1}^K \sum_{j=1}^K p_{ij} \log\frac{p_{ij}}{u_{ij}}\,.
}
\end{equation}

Minimizing this loss forces the appearance model to learn appearance features that agree with the learned associations. The rationale to choose appearance pairs that are temporally $T$ frames apart is to capture appearance changes over a longer period instead of incremental changes between two frames.  

\noindent{\textbf{Preprocessing Detections.}}
In reality, objects frequently enter and exit the scene, and false positive and negative detections may occur. We therefore preprocess the detections such that each training batch contains exactly $K$ objects throughout $T$, although an explicit object birth and death handling can be adopted~\cite{miah2025learning}.

\noindent{\textbf{Features.}} 
Given a pair of detections $\mathbf{z}_i = (x_i, y_i, w_i, h_i)$ 
and $\mathbf{z}_j = (x_j, y_j, w_j, h_j)$, where $x$, $y$ $w$, $h$ indicate bounding box's central coordinate and its width and height, the pairwise geometric feature is computed as: $\mathbf{f}_{ij} = \left(\frac{2(x_j - x_i)}{h_i + h_j}, \frac{2(y_j - y_i)}{h_i + h_j}, \log \frac{h_i}{h_j}, \log \frac{w_i}{w_j}, \text{IoU}\right)$. These pairwise features serve as the input to $g_{\theta}(\mathord{\cdot})$ for regressing the score matrix $\mathbf{S}$. The network $g_{\theta}(\mathord{\cdot})$ is implemented as two-layer MLP with ReLU non-linearity and is trained end-to-end using stochastic gradient descent. We provide further implementation and training details in the supplementary material.

\subsection{Inference}\label{subsec:inference}
For testing, we utilize Tracktor~\cite{bergmann2019tracking} to preprocess the provided raw detections as suggested in~\cite{hornakova2020lifted,bastani2021self,lu2024self}. We use the learned network to associate detections with predicted objects using a Kalman filter. Following~\cite{zhang2022bytetrack}, we define state $\mathbf{x} = (x, y, w, h, \dot{x}, \dot{y}, \dot{w}, \dot{h})$ to denote bounding box central coordinates and its corresponding velocities. More details regarding the process and observation noise of the Kalman Filter are provided in the supplementary material.

\noindent{\textbf{Combining Motion and Appearance Cues.}} 
We use the learned $g_{\theta}$ and $\phi_{\theta}$ to associate detections with the predicted objects. Suppose at frame $t$, we have $N$ bounding boxes $\hat{\mathbf{z}}_t$ predicted by the motion model along with $M$ detections $\mathbf{z}_t$, the cost matrix $\mathbf{C} \in \mathbb{R}^{N \times M}$ is then used for association:
%
\begin{equation}\label{eq:app_cost}
c_{ij} = -g_{\theta}(\hat{\mathbf{z}}^i_t, \mathbf{z}^j_t) - \kappa \left(\frac{\phi_{\theta}(\hat{\mathbf{z}}^i_t)^T\phi_{\theta}(\mathbf{z}^j_t)}{||\phi_{\theta}(\hat{\mathbf{z}}^i_t)||^2||\phi_{\theta}(\mathbf{z}^j_t)||^2} - s_{\text{min}}\right),
\end{equation}
where $s_{\min}$ is the minimum cosine similarity threshold for two detections belonging to the same track and $\kappa$ is the scaling factor for the cosine similarity between two detections. 

\noindent{\textbf{Detection Noise Handling.}}
In practice, a predicted box at frame $t$ might not be matched to any detection due to occlusion or a missing detection. Vice versa, a detection at frame $t$ may not be matched to any of the predicted boxes as it could be a false positive or start of a new track. To handle this, we propose to augment $\mathbf{C}$ with an auxiliary row and column containing a learned cost $c_{\text{miss}}$ for a missing association, such that $\mathbf{C} \in \mathbb{R}^{(N+1) \times (M+1)}$. We then obtain the optimal data association $\mathbf{A}^\ast$ by solving 
%
%
%
\begin{equation}\label{eq:lap}
\scalemath{0.85}{
\displaystyle \mathbf{A}^\ast = \mathop{\arg \min}_{\mathbf{A} \in \mathcal{A}} \sum_{i=1}^{N+1} \sum_{j=1}^{M+1} c_{ij} a_{ij} \quad \text{s.t.}\ \sum_{j=1}^{M+1} a_{ij} = 1, \forall i \in \{1, \cdots, N + 1\}, \\
\sum_{i=1}^{N+1} a_{ij} = 1, \forall j \in \{1, \cdots ,M + 1\}
}
\end{equation}
using the Hungarian algorithm~\cite{kuhn1955hungarian}. In this way, tracks that are unmatched will only be updated by the motion model and detections that are unmatched to predictions initialize a new track if the detection confidence is high enough, so that our approach can deal with newly entering objects. Tracks that remain unmatched for more than $\tau$ frames are terminated.

\vspace{-0.3cm}
\section{Experiments}
\subsection{Datasets}
Our experiments are conducted on the MOT17/20 and BDD100K~\cite{yu2020bdd100k} datasets. MOT17 contains 7 videos for training and 7 for testing. For all videos, detections from DPM~\cite{felzenszwalb2009object}, FRCNN~\cite{ren2015faster} and SDP~\cite{yang2016exploit} are provided, resulting in 21 videos in total for training and testing. For MOT20, 4 videos are provided for training and testing under crowded scenarios. We report CLEAR~\cite{bernardin2008evaluating} metrics such as MOTA and number of identity switches (IDSW), IDF1 and the HOTA metric~\cite{luiten2021hota}. BDD100K~\cite{yu2020bdd100k} is a large-scale autonomous driving dataset that has 1400 and 200 videos in the training and validation set, respectively. The videos contain fast camera ego-motion and frequent occlusions. 8 classes including cars and pedestrians are included. mHOTA and mIDF1 are averaged across all classes, whereas IDF1 and IDSW are summed over all classes.

\vspace{-0.3cm}
\subsection{Ablation Study}
\begin{table}[!t]
\centering
\scriptsize
\tabcolsep=0.75cm
\begin{tabular}{c c c c c} 
 \hline
Association  & HOTA $\uparrow$ & MOTA $\uparrow$ & IDF1 $\uparrow$ & IDSW $\downarrow$ \\
 \hline
mot & 62.0 & 64.0 & 69.9 & 731\\
mot + app & \textbf{62.4} & \textbf{64.1} & \textbf{70.5} & \textbf{652}\\
 \hline
\end{tabular}
\caption{Results on the MOT17 training set using \textbf{public} detections.}
\label{tab:mot17ablationreid}
\vspace{-3mm}
\end{table}

\begin{figure}[!t]
  \centering
  \begin{minipage}[b]{0.45\linewidth}
    \centering
    \includegraphics[width=\linewidth]{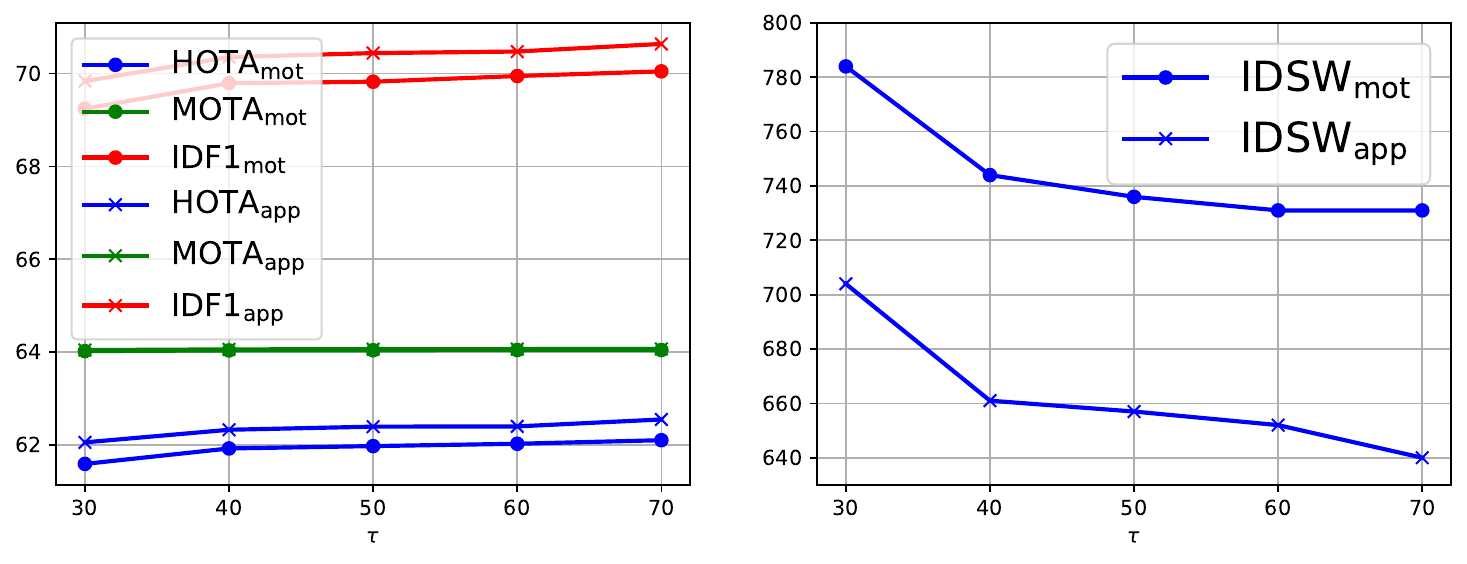}\\[\smallskipamount]
    \footnotesize
    \vspace{-0.2cm}
    a) Tracking with different values of $\tau$
    \label{figure:tau}
  \end{minipage}
  \quad
  \begin{minipage}[b]{0.45\linewidth}
    \centering
    \includegraphics[width=\linewidth]{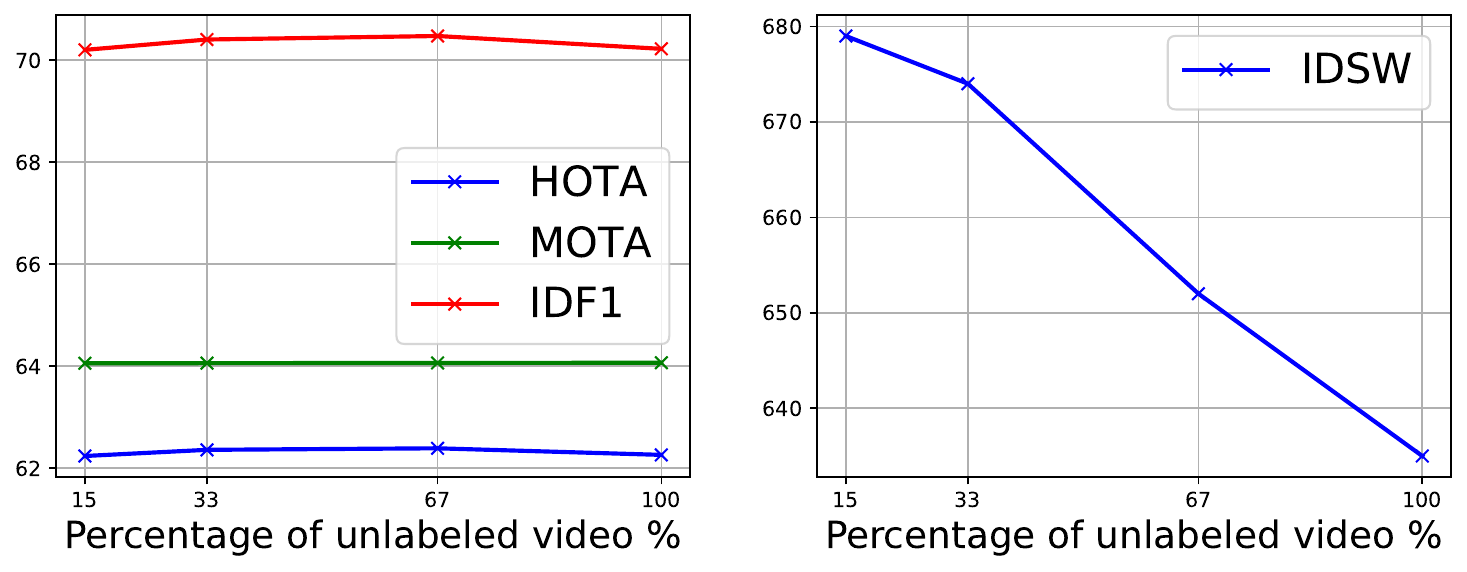}\\[\smallskipamount]
     \footnotesize
     \vspace{-0.25cm}
    b) Varying the percentage of unlabeled videos
    \label{figure:scaling_exp}
  \end{minipage}
  \vspace{-0.2cm}
  \caption{Ablation studies on MOT17 training set.}
  \label{figure:ablation}
  \vspace{-0.3cm}
\end{figure}

\noindent{\textbf{Impact of Appearance Cost.}}
To show the effectiveness of fine-tuning the appearance model, we conduct experiments on the MOT17 training set. As can be seen in~\Cref{tab:mot17ablationreid}, adding appearance cost can further boost the model's tracking performance on all metrics. In particular, it reduces the number of identity switches (IDSW) by 11\%.

\noindent{\textbf{Impact of $\tau$.}}
We study how the tracking performance is affected by the number of frames used for re-identification after occlusion. From Fig.~\ref{figure:ablation}(a), we observe a consistent gain among tracking metrics by increasing $\tau$. By default, we use $\tau = 60$. 

\noindent{\textbf{Impact of Amount of Unlabeled Data.}}
We study how the tracking performance varies when training with different amount of unlabeled videos. We vary the training corpus by using 100\%, 67\%, 33\%, and 15\% of MOT17 and report the results on the training set with public detections. Results are shown in Fig.~\ref{figure:ablation}(b). While the difference in HOTA, MOTA, and IDF1 metrics is small, we observe that IDSW decreases with more training data, indicating the improved association with more unlabeled training videos.

\noindent{\textbf{Impact of Number of Sinkhorn Iterations.}}
Our method is not sensitive to the number of Sinkhorn iterations as long as it exceeds 5. We used 20 iterations in our experiments. 

\begin{table}[!t]
     \centering
    \scriptsize
    \tabcolsep=0.6cm
    \begin{tabular}{c c c c c c}
    \toprule
    Method & Sup. & HOTA $\uparrow$ & MOTA $\uparrow$ & IDF1 $\uparrow$ & IDSW $\downarrow$\\
    \midrule
    MHT\_BiLSTM~\cite{kim2018multi}        & \Checkmark & 41.0 & 47.5 & 51.9 & 2069\\
    Tracktor++~\cite{bergmann2019tracking} & \Checkmark & 44.8 & 56.3 & 55.1 & 1987\\
    TrackFormer~\cite{meinhardt2022trackformer} & \Checkmark & - & 57.6 & 62.3 & 4018\\
    SUSHI~\cite{cetintas2023unifying}      & \Checkmark & 54.6 & 62.0 & 71.5 & 1041\\
    \midrule
    SORT~\cite{bewley2016simple}           & \XSolidBrush & - & 43.1 & 39.8 & 4852\\
    UNS20regress~\cite{bastani2021self}    & \XSolidBrush & 46.4 & 56.8 & 58.3 & 1320\\
    UnsupTrack~\cite{karthik2020simple} & \XSolidBrush & 46.9 & \textbf{61.7} & 58.1 & 1864\\
    Lu~\etal~\cite{lu2024self}                   & \XSolidBrush & \underline{49.0} & 58.8 & \underline{61.2} & \textbf{1219}\\
    Ours                    & \XSolidBrush & \textbf{50.3} & \underline{60.3} & \textbf{63.4} & \underline{1266}\\
    \bottomrule
    \end{tabular}
    \vspace{-0.4cm}
    \caption{Benchmark results on MOT17 test set using \textbf{public} detections, \Checkmark indicates fully-supervised and \XSolidBrush means self-supervised methods. The best and second best self-supervised performances are shown in bold and underlined numbers, respectively}
    \label{tab:motchallenge_mot17_public}
\end{table}

\begin{table}[!t]
     \centering
    \scriptsize
    \tabcolsep=0.6cm
    \begin{tabular}{c c c c c c}
    \toprule
    Method & Sup. & HOTA $\uparrow$ & MOTA $\uparrow$ & IDF1 $\uparrow$ & IDSW $\downarrow$\\
    \midrule
    Tracktor++V2~\cite{bergmann2019tracking}& \Checkmark & 42.1 & 52.6 & 52.7 & 1648\\
    ArTist~\cite{saleh2021probabilistic}& \Checkmark & - & 53.6 & 51.0 & 1531\\
    ApLift~\cite{hornakova2021making} & \Checkmark & 46.6 & 58.9 & 56.5 & 2241\\
    SUSHI~\cite{cetintas2023unifying} & \Checkmark & 55.4 & 61.6 & 71.6 & 1053\\
    \midrule
    SORT20~\cite{bewley2016simple} & \XSolidBrush & 36.1 & 42.7 & 45.1 & 4470\\ 
    Ho~\etal~\cite{ho2020two} & \XSolidBrush & - & 41.8 & - & 5918\\
    UnsupTrack~\cite{karthik2020simple} & \XSolidBrush & \underline{41.7} & \underline{53.6} & \underline{50.6} & \underline{2178}\\
    Ours & \XSolidBrush & \textbf{47.4} & \textbf{59.5} & \textbf{58.5} & \textbf{1656}\\
    \bottomrule
    \end{tabular}
    \caption{Benchmark results on MOT20 test set using \textbf{public} detections, \Checkmark indicates fully-supervised and \XSolidBrush means self-supervised methods. The best and second best self-supervised performances are shown in bold and underlined numbers, respectively}
    \label{tab:motchallenge_mot20}
\end{table}

\begin{table}[!t]
    \centering
    \scriptsize
    \tabcolsep=0.6cm
    \begin{tabular}{c c c c c c}
    \toprule
    Method & Sup. & mHOTA $\uparrow$ & mIDF1 $\uparrow$ & IDF1 $\uparrow$ & IDSW $\downarrow$\\
    \midrule
    MOTR~\etal~\cite{zeng2022motr} & \Checkmark & - & 43.5 & - & -\\
    Yu~\etal~\cite{yu2020bdd100k} & \Checkmark & - & 44.5 & 66.8 & 8315\\
    QDTrack~\cite{pang2021quasi} & \Checkmark & 41.7 & 50.8 & 71.5 & 6262\\
    ByteTrack~\cite{zhang2022bytetrack} & \Checkmark & - & 54.8 & 70.4 & 9140\\
    \midrule
    SORT~\cite{bewley2016simple} & \XSolidBrush & 27.9 & 33.8 & 56.4 & \textbf{9647}\\ 
    Ours & \XSolidBrush & \textbf{34.5} & \textbf{42.2} & \textbf{62.3} & 23143\\
    \bottomrule
    \end{tabular}
    \caption{Benchmark results on BDD100K~\cite{yu2020bdd100k} validation set. The best self-supervised performances are shown in bold numbers. We use the same detections as in~\cite{zhang2022bytetrack}.}
    \label{tab:bdd100k}
\end{table}

\subsection{Comparison with State of the Art}

\noindent{\textbf{MOT17.}} We compare our method with several other approaches in~\Cref{tab:motchallenge_mot17_public}. SORT~\cite{bewley2016simple} relies on heuristic IoU matching with a Kalman filter. UnsupTrack~\cite{karthik2020simple} uses SORT to generate pseudo labels for learning an appearance model. UNS20regress~\cite{bastani2021self} utilizes motion and appearance consistency trained with an RNN for association. Lu \etal~\cite{lu2024self} impose a heuristic path consistency constraint with several loss terms to train a matching network. Our method uses exactly the same input detections with~\cite{bastani2021self,lu2024self} preprocessed with Tracktor, and outperforms their approaches in terms of HOTA and IDF1. UnsupTrack~\cite{karthik2020simple} achieves better MOTA, but it uses a better detector, \ie\ CenterNet, to improve public detections.

It is worth noting that our method performs even better than several fully-supervised approaches like MHT\_BiLSTM~\cite{kim2018multi}, TrackFormer~\cite{meinhardt2022trackformer}. Tracking results using private detections are provided in the supplementary material.

\noindent{\textbf{MOT20.}} \Cref{tab:motchallenge_mot20} shows our results. Our method outperforms the strongest self-supervised baseline UnsupTrack~\cite{karthik2020simple} in all tracking metrics by a large margin. 

\noindent{\textbf{BDD100K.}} We train our model on the BDD100K training set and report the tracking results on the validation set in~\Cref{tab:bdd100k}, using the same YOLOX detector as in~\cite{zhang2022bytetrack}. For completeness, we evaluated SORT~\cite{bewley2016simple} as well. Our approach outperforms SORT for all metrics, only IDSW is lower for SORT. This, however, can be explained by the low recall of SORT ($26.0$ IDentity Recall). SORT therefore tracks only the simple cases and misses many tracks, which results in a low IDSW. Our approach has a much higher recall ($36.5$ IDR), which comes at the cost of more IDSWs. The results show that our approach also performs well on large-scale MOT datasets.

\begin{figure}[!t]
  \centering
  \includegraphics[width=\textwidth]{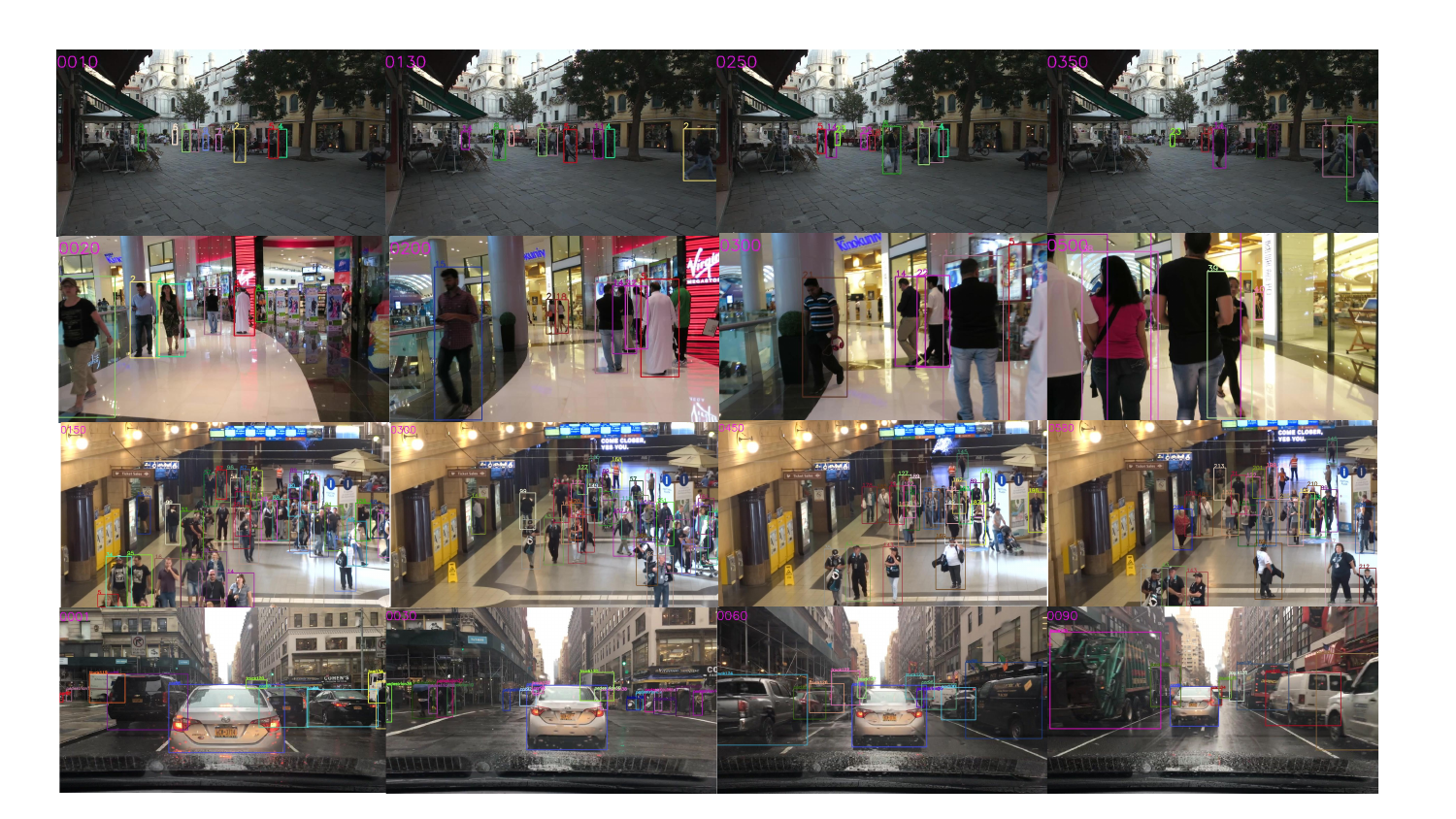}
  \vspace{-0.8cm}
	\caption{Qualitative results of our tracking method on the MOT17/20 and BDD100K dataset. Our method is able to track objects of different classes under occlusions, camera ego-motion and crowded scenarios. Best viewed in color.}
 \label{figure:qualitative_results}
\end{figure}

\noindent{\textbf{Qualitative Results.}} \Cref{figure:qualitative_results} shows some qualitative results for the MOT17/20 and BDD100K test set. Our method can track objects with similar appearances under occlusion, camera motion and crowded scenarios, despite learning data associations in a self-supervised manner.

\section{Conclusion}
In this work, we introduced a maximum likelihood learning framework for self-supervised multi-object tracking. It learns a network for associating detections between adjacent frames and an appearance model by associating detections to states of a Kalman filter. Our method does not require expensive identity-level annotations for training, it enables online inference and achieves state-of-the-art performances on the MOT17 and MOT20 datasets among existing self-supervised approaches given public detections. It also achieves promising results on the large-scale BDD100K dataset. 

\section*{Acknowledgements} The work has been supported by the project iBehave (receiving funding from the programme ``Netzwerke 2021'', an initiative of the Ministry of Culture and Science of the State of Northrhine Westphalia) and the ERC Consolidator Grant FORHUE (101044724). Shuai would like to thank Andreas Döring for technical discussions and Sicong Pan for proofreading the draft. The sole responsibility for the content of this publication lies with the authors. 
\vspace{-0.8cm}

\section*{Appendix}

\section{Implementation Details}
\noindent{\textbf{Preprocessing Detections.}}
During training, we use Faster RCNN~\cite{ren2015faster} detections, and our training formulation attempts to track $K$ objects in a single clip, where $K$ can vary among clips.
We threshold the detections based on the detection confidence values to remove potential false positives. The number of remaining detections in the first frame of each clip defines $K$. To compensate for missing detections, 
we use the KCF~\cite{henriques2014high} tracker, initialized independently for each detection at the first frame of the clip. In case of missing detections at certain frames, we use the bounding box output of KCF. If this results in more than $K$ detections in the next frame, we discard detections with a low intersection-over-union with tracked bounding boxes. 
This preprocessing operation enables us to collect $K \times T$ detections from a single video clip, where $T$ is the clip length. While $K$ can differ for each clip, we keep $T$ the same for all clips. For the sake of computational efficiency, we generate the video clips with $T=10$ from each training sequence. Overall, we generate 260 videos from the MOT17 training set to train the association network that is used for tracking on MOT17 and MOT20, the same preprocessing procedure is used to train matching network on BDD100K.

\noindent{\textbf{Training.}}
Our implementation is based on the PyTorch framework. We use the detections' first frame coordinates to initialize mean $\boldsymbol{\mu}_1$ of the first state $\mathbf{x}_1$, and the covariance matrix $\boldsymbol{\Sigma}_1$ is initialized as a diagonal matrix with variance set to 300. For the motion model $\mathbf{F}$ in the Kalman filter, a constant velocity model is utilized. The process noise $\sigma_q$ in the diagonal covariance matrix $\mathbf{Q}$ is set to 150 and the observation noise $\sigma_r$ in the diagonal covariance matrix $\mathbf{R}$ is set to 5. This enables the Kalman filter to rely more on the observation model during training. We have also experimented with updating the parameters of $\mathbf{Q}$ during training but the differences were marginal as long as $\mathbf{Q}$ is initialized to a large value. 
For each bounding box $\mathbf{z}$, we resize it to $224 \times 224$ and feed it into $\phi_{\theta}$, which is parametrized by ImageNet pretrained ResNet-50~\cite{he2016deep}, to obtain the appearance embedding $\phi_{\theta}(\mathbf{z})$, followed by $L_2$ normalization. Adam~\cite{kingma2014adam} optimizer is used during training. We train $g_{\theta}(\mathord{\cdot})$ with a learning rate of $5\times10^{-3}$ for 10 epochs followed by fine-tuning  $\phi_{\theta}$ with a learning rate of $10^{-4}$ for another 3 epochs. Note that we only use the \textit{unlabeled} detections from the MOT17 training set to fine-tune $\phi_{\theta}$.

\vspace{-0.58cm}
\section{Additional Ablation Studies}
\label{sec:A}

\begin{table}[!t]
    \centering
    \scriptsize
    \tabcolsep=0.9cm
    \scalebox{0.9}{
    \begin{tabular}{r c c c c}
    \toprule
    $s_{\min}$ & HOTA $\uparrow$ & MOTA $\uparrow$ & IDF1 $\uparrow$ & IDSW $\downarrow$\\
    \midrule
    \parbox[t]{5mm}{\multirow{4}{*}{\rotatebox[origin=c]{90}{$\kappa=5$}}}
    0.7 & 61.2 & 64.0 & 68.8 & 712\\
    0.75 & 61.9 & 64.0 & 70.0 & 690\\
    0.8 & 62.2 & 64.1 & 70.3 & 668\\
    0.85 & 62.4 & \textbf{64.1} & 70.5 & \textbf{652}\\
    0.9 & \textbf{62.5} & 64.1 & \textbf{70.6} & 656\\
    \bottomrule
    \end{tabular}
    }
\end{table}

\begin{table}[!t]
    \centering
    \scriptsize
    \tabcolsep=0.9cm
    \scalebox{0.9}{
    \begin{tabular}{r c c c c c}
    \toprule
    $s_{\min}$ & HOTA $\uparrow$ & MOTA $\uparrow$ & IDF1 $\uparrow$ & IDSW $\downarrow$\\
    \midrule
      \parbox[t]{5mm}{\multirow{4}{*}{\rotatebox[origin=c]{90}{$\kappa=10$}}}
    0.7 & 59.9 & 64.0 & 66.0 & 697\\
    0.75 & 60.8 & 64.0 & 67.6 & 669\\
    0.8 & 61.8 & 64.0 & 69.4 & \textbf{647}\\
    0.85 & \textbf{62.3} & \textbf{64.1} & \textbf{70.4} & 649\\
    0.9 & 62.3 & 64.1 & 70.2 & 653\\
    \bottomrule
    \end{tabular}
    }
    \caption{Tracking performance under different combinations of hyperparameters on the MOT-17 training set.}
    \label{tab:motchallenge_mot17_app_ablation}
\end{table}

We study the influence of the parameters $\kappa$ and  $s_{\min}$ (11) for the tracking performance on the MOT17 training set. The results are shown in~\Cref{tab:motchallenge_mot17_app_ablation}. The results show that our approach is not very sensitive to the parameters. As default values, we use $\kappa=5$ and $s_{\min}=0.85$.

\vspace{-0.5cm}
\section{Effect of Kalman Filter's hyper-parameters}
\label{sec:B}

During inference, we initialize the mean of each track as: $\boldsymbol{\mu}_{\text{init}}=(x, y, w, h, 0, 0, 0, 0)$ where $x, y,w,h$ denotes the bound-box's center and width/height. The initial covariance is dependent on the specific detection, \ie, \scalebox{0.58}{$\boldsymbol{\Sigma}_{\text{init}}=\text{diag}((2\sigma_{\text{pos}}w)^2, (2\sigma_{\text{pos}}h)^2, (2\sigma_{\text{pos}}w)^2, (2\sigma_{\text{pos}}h)^2, (10\sigma_{\text{vel}}w)^2, (10\sigma_{\text{vel}}h)^2, (10\sigma_{\text{vel}}w)^2, (10\sigma_{\text{vel}}h)^2)$} where $\sigma_{\text{pos}}$ and $\sigma_{\text{vel}}$ are the variance for bounding box's position and velocity, respectively. For process covariance: \scalebox{0.68}{$\mathbf{Q}=\text{diag}((\sigma_{\text{pos}}w)^2,(\sigma_{\text{pos}}h)^2,(\sigma_{\text{pos}}w)^2,(\sigma_{\text{pos}}h)^2,(\sigma_{\text{vel}}w)^2,(\sigma_{\text{vel}}h)^2,(\sigma_{\text{vel}}w)^2,(\sigma_{\text{vel}}h)^2)$} and for measurement noise: \scalebox{0.62}{$\mathbf{R} = \text{diag}((\sigma_{\text{pos}}w)^2, (\sigma_{\text{pos}}h)^2, (\sigma_{\text{pos}}w)^2, (\sigma_{\text{pos}}h)^2)$}.

We study the influence of these hyper-parameters on the performance on the MOT17 training set using public detections. We only use the motion affinity network during tracking. The results in Table~\ref{tab:kalman_noise_ablation} suggest that the tracking performance is not very sensitive to the choice of process and measurement noise. Therefore, we set $\sigma_{\text{pos}}=\frac{1}{20}$ and $\sigma_{\text{vel}}=\frac{1}{160}$ in the final model during inference. In particular, we use the same noise parameters of the Kalman filter for MOT17, MOT20 and BDD100K~\cite{yu2020bdd100k}. 

We also provide inference time of our approach in~\Cref{tab:inference_fps}. Thanks to our online inference procedure, our approach is fast and can be used in real-time applications.

\begin{table}[!t]
\centering
\scriptsize
\tabcolsep=0.5cm
\begin{tabular}{c|c|c|c} 
 \hline
 & HOTA$\uparrow$/IDF1$\uparrow$/IDSW$\downarrow$ & HOTA$\uparrow$/IDF1$\uparrow$/IDSW$\downarrow$  &  HOTA$\uparrow$/IDF1$\uparrow$/IDSW$\downarrow$ \\
 \hline
  & $\sigma_{\text{vel}}=\frac{1}{320}$ & $\sigma_{\text{vel}}=\frac{1}{160}$  & $\sigma_{\text{vel}}=\frac{1}{80}$ \\
 \hline
 $\sigma_{\text{pos}}=\frac{1}{40}$ & 62.0/69.9/731 & 62.0/69.8/725 & 61.4/69.0/783  \\
 \hline
 $\sigma_{\text{pos}}=\frac{1}{20}$ & 62.0/69.9/727 & 62.0/69.9/731 & 62.0/69.8/725  \\ 
 \hline
 $\sigma_{\text{pos}}=\frac{1}{10}$ & 61.8/69.7/773 & 62.0/69.9/727 & 62.0/69.8/731  \\
 \hline
 $\sigma_{\text{pos}}=\frac{1}{5}$ & 61.4/69.2/834 & 61.8/69.7/773 & 62.0/69.9/727 \\
 \hline
\end{tabular}
\caption{Impact of different parameters for the Kalman filter on the MOT17 training set with public detections. The results are reported without appearance model.}
\label{tab:kalman_noise_ablation}
\end{table}

\begin{table}[!t]
\centering
\scriptsize
\tabcolsep=1.6cm
\begin{tabular}{c c c} 
 \hline
 Association & MOT17 & MOT20  \\
 \hline
 mot & 96 & 12 \\
 mot+app & 33 & 6 \\
 \hline
\end{tabular}
\caption{Inference speed (FPS) on different datasets.}
\label{tab:inference_fps}
\end{table}

\begin{table}[!t]
     \centering
    \scriptsize
    \tabcolsep=0.6cm
    \begin{tabular}{c c c c c c}
    \toprule
    Method & Sup. & HOTA $\uparrow$ & MOTA $\uparrow$ & IDF1 $\uparrow$ & IDSW $\downarrow$\\
    \midrule
    MOTR~\cite{zeng2022motr} & \Checkmark & 57.8 & 73.4 & 68.6 & 2439\\
    MeMOTR~\cite{gao2023memotr} & \Checkmark & 58.8 & 72.8 & 71.5 & 1902\\
    MOTRv2~\cite{zhang2023motrv2} & \Checkmark & 62.0 & 78.6 & 75.0 & 2619\\
    \midrule
    UCSL~\cite{meng2023tracking} & \XSolidBrush & 58.4 & 73.0 & 70.4 & -\\
    OUTrack~\cite{liu2022online} & \XSolidBrush & 58.7 & 73.5 & 70.2 & 4122\\
    ByteTrack~\cite{zhang2022bytetrack} & \XSolidBrush & 63.1 & \underline{80.3} & 77.3 & 2196\\
    U2MOT~\cite{liu2023uncertainty} & \XSolidBrush & \underline{64.2} & 79.9 & \underline{78.2} & \underline{1506}\\
    Lu~\etal~\cite{lu2024self} & \XSolidBrush & \textbf{65.0} & \textbf{80.9} & \textbf{79.6} & 1749\\
    Ours & \XSolidBrush & 62.1 & 76.7 & 75.7 & \textbf{1092}\\
    \bottomrule
    \end{tabular}
    \caption{Benchmark results on MOT17 test set using \textbf{private} detections, \Checkmark indicates fully-supervised and \XSolidBrush means self-supervised methods. The best and second best self-supervised performances are shown in bold and underlined numbers, respectively}
    \label{tab:motchallenge_mot17_private}
\end{table}

\section{Results on MOT17 using Private Detections} 
We also compare our approach with other works under the private detection protocol in~\Cref{tab:motchallenge_mot17_private}. We use the same YOLOX detector as~\cite{zhang2022bytetrack}. Results show that our tracker achieves competitive results with the state of the art~\cite{liu2023uncertainty, lu2024self} and even outperforms several transformer-based methods~\cite{zeng2022motr,zhang2023motrv2,gao2023memotr} that require expensive identity-level supervision and have a much larger number of parameters. 

\bibliography{egbib}
\end{document}